# Hybrid Mamba-Transformer Network for Single-Shot 3D Reconstruction from Fringe Patterns

Xinjie Zhou *a, XiaoLei Li**b
a. Aulin College, Northeast Forestry University, Harbin,150040, China
b.School of Mechanical Engineering, Hebei University of Technology, Tianjin, 300130, China
*Corresponding author: zhouxinjie@nefu.edu.cn;
**Corresponding author: lixiaolei_hebut@163.com

**Abstract:** Deep learning facilitates rapid single-shot three-dimensional (3D) reconstruction using fringe projection profilometry (FPP), but recovering accurate surface height from a single high-frequency fringe image remains challenging because both long-range spatial consistency and fine geometric details must be preserved. Convolutional networks are limited in modeling long-range dependencies, whereas window-based Transformers establish global interactions only progressively with increasing depth. To address these limitations, a hybrid Mamba–Transformer network is proposed for fringe-to-height regression. The network adopts a Swin-based U-shaped architecture and introduces visual state-space modeling at the intermediate-resolution encoder stage. An axial–dilated vision state-space (ADVSS) block is further developed to enhance pre-scan feature encoding through local, dilated, and axial depth-wise convolutions while retaining four-directional selective scanning. This design combines linear-complexity long-range modeling with direction-aware preservation of local fringe structures. Experiments on a publicly available FPP dataset show that the proposed method outperforms hNet, SwinConvUNet, UHRNet, and HiDNet in mean absolute error(MAE), root-mean-square error (RMSE), absolute relative error, and structural similarity index measure (SSIM). Using only one deformed fringe image, the proposed method achieves an RMSE of 0.548 mm and an SSIM of 0.9985, reducing RMSE by 65.2% and 59.6% relative to UHRNet and HiDNet, respectively. These results demonstrate the effectiveness of combining selective state-space modeling with hierarchical window-based attention for accurate single-shot FPP reconstruction.

## 1. Introduction

Fringe projection profilometry (FPP) is a well-established optical metrology technique for noncontact, high-precision three-dimensional (3D) surface measurement [1–3]. In a typical FPP system, sinusoidal fringe patterns are projected onto the measured object, and the phase distortions caused by the surface topography are analyzed to recover depth through geometric triangulation. Conventional FPP methods achieve high measurement accuracy using multistep phase shifting and temporal phase unwrapping [4]. However, the acquisition of multiple sequential fringe images makes these methods susceptible to motion-induced errors and limits their applicability to dynamic scenes and real-time measurements.

Single-shot FPP has been developed to overcome this limitation by recovering 3D geometry from only one fringe image. Traditional single-shot approaches, including Fourier transform profilometry (FTP) [5] and other spatial phase-demodulation techniques, are often affected by spectral leakage, frequency aliasing, and phase errors near discontinuities and geometrically complex regions. In recent years, deep learning has provided an alternative solution by learning a direct mapping from a single fringe pattern to phase or depth [6–11], thereby reducing or eliminating the need for explicit phase demodulation and unwrapping. Most existing methods rely on convolutional neural networks (CNNs), which are effective at extracting local spatial features. Nevertheless, convolution is inherently a local operation, and its effective receptive field expands only gradually with network depth. As high-frequency fringe patterns contain periodic structures extending across the entire field of view, insufficient long-range interaction may lead to inconsistencies in depth recovery, particularly in regions containing discontinuities, steep slopes, shadows, or low fringe modulation.

Transformers [12] provide a potential solution by modeling long-range dependencies through self-attention. However, the quadratic computational and memory complexity of global self-attention with respect to the number of image tokens restricts its application to high-resolution optical images. The Swin Transformer [13] alleviates this problem by alternately applying window-based and shifted-window multi-head self-attention (W-MSA and SW-MSA), resulting in approximately linear computational complexity while enabling information exchange across neighboring windows. Swin-based U-Net architectures have subsequently been introduced into single-shot structured-light profilometry and fringe-phase retrieval [14,15], demonstrating improved contextual modeling compared with purely convolutional networks. Despite these advantages, interactions between distant image regions must still be established progressively through multiple shifted-window blocks. Consequently, the effective receptive field may remain insufficient for maintaining image-wide consistency in high-resolution fringe images. Moreover, conventional square attention windows do not explicitly exploit the pronounced directional and periodic characteristics of projected fringe patterns.

State-space models (SSMs) [16] have recently emerged as computationally efficient alternatives for long-range dependency modeling. Mamba combines an input-dependent selective mechanism with linear-complexity sequence scanning. Its vision-oriented variants, such as VMamba, Vision Mamba (Vim), and VM-UNet [17–19], extend this mechanism to images through bidirectional or multidirectional scanning. These characteristics make visual SSMs particularly attractive for modeling the long-range continuity of fringe patterns. However, directly employing a pure Mamba encoder may not fully satisfy the requirements of high-precision fringe-to-depth regression. Recovering metric depth from a single high-frequency fringe image, for example, one containing 128 vertical fringes, requires both long-range consistency and accurate preservation of local geometric details. Although multidirectional selective scanning can efficiently propagate global information, it may insufficiently emphasize fine structures around object boundaries, steep surface variations, shadowed areas, and low-modulation regions. Conversely, Swin Transformer blocks provide effective hierarchical feature refinement but require multiple stages of shifted-window propagation to establish global relationships. Therefore, neither a purely window-based Transformer nor an entirely SSM-based architecture provides an optimal balance between global dependency modeling and local-detail preservation for this task.

Hybrid Mamba – Transformer architectures have recently demonstrated that selective state-space modeling and attention can provide complementary global and local representations [20–22]. Motivated by this observation, this study proposes a hybrid Mamba–Transformer network for single-shot fringe-to-depth reconstruction. The proposed network retains a Swin-based U-Net architecture and incorporates visual state-space (VSS) blocks into intermediate encoder stages. At these stages, the reduced spatial resolution offers a favorable balance between computational efficiency and long-range feature modeling. The VSS blocks employ linear-complexity multidirectional selective scanning to propagate fringe information over extended spatial ranges and improve global consistency. Meanwhile, the Swin-based components perform hierarchical local aggregation and facilitate the preservation and reconstruction of fine geometric details.

To further adapt selective state-space modeling to the characteristics of fringe images, an axial- and dilation-aware visual state-space (ADVSS) module is developed. Before multidirectional selective scanning, axial convolutions enhance orientation-dependent features along the horizontal and vertical directions, while dilated convolutions enlarge the local receptive field and capture multiscale spatial structures. This front-end enhances the directional and periodic features of high-frequency fringes without modifying the underlying selective-scan mechanism. Consequently, the proposed architecture combines the efficient global modeling capability of Mamba with the local inductive bias and hierarchical feature representation of the Swin Transformer.

The main contributions of this study are summarized as follows:

1.A hybrid Mamba–Transformer architecture is proposed for direct depth reconstruction from a single high-frequency fringe image. VSS blocks are incorporated into selected encoder stages of a Swin-based U-Net to balance long-range dependency modeling, local-detail preservation, and computational cost.

2.An ADVSS module is developed to adapt visual state-space modeling to the directional and periodic characteristics of fringe patterns. Axial and dilated convolutions enhance orientation-sensitive and multi-scale local features before multi-directional selective scanning.

The proposed method is comprehensively evaluated against hNet, SwinConvUNet, UHRNet, and HiDNet under a unified training and evaluation protocol. The experiments include quantitative comparisons, qualitative evaluations, and ablation studies to assess the effectiveness of the hybrid architecture and the ADVSS module.

3.For a fair comparison, all baseline networks are reimplemented using the same training protocol and evaluated on a common dataset constructed from the publicly available fringe-projection data reported by Wang et al. [7,8]. The proposed method consistently outperforms the compared approaches in terms of mean absolute error (MAE), root-mean-square error (RMSE), absolute relative error (AbsRel), and structural similarity index measure (SSIM). It achieves an RMSE of 0.548 mm and an SSIM of 0.9985, corresponding to RMSE reductions of 65.2% and 59.6% relative to UHRNet and HiDNet, respectively. These results demonstrate the effectiveness of integrating selective state-space modeling with hierarchical window-based attention for high-precision single-shot FPP.

The remainder of this paper is organized as follows. Section 2 presents the proposed network architecture and loss formulation. Section 3 describes the experimental setup and reports the comparative and ablation results. Section 4 summarizes the main findings and discusses directions for future research.

## 2. Method

The proposed framework consists of two stages: (i) offline training of the hybrid Mamba–Transformer network for fringe-to-depth regression and (ii) online inference for reconstructing a metric depth map from a single captured fringe pattern. The overall framework is illustrated in Fig.

1. Specifically, Fig. 1(a) presents the offline training procedure, whereas Fig. 1(b) shows the online 3D reconstruction process for a target object.

During offline training, as shown in Fig. 1(a), fringe patterns from the training dataset are fed into the proposed network to generate the corresponding depth maps. The discrepancy between each predicted depth map and its ground-truth counterpart is quantified using a loss function. The network parameters are then iteratively updated by an optimizer to minimize the training loss. During online inference, as illustrated in Fig. 1(b), a single fringe image captured by the camera of the FPP system is processed by the trained network. The corresponding metric depth map is generated in a feed-forward manner and subsequently used to recover the 3D surface of the target object.

The proposed network is designed to improve the accuracy of single-shot fringe-to-depth regression by jointly modeling long-range fringe continuity and fine local geometry. To achieve this objective, visual state-space (VSS) blocks are incorporated into the intermediate encoder stages of a Swin-based U-Net. At these stages, multi-directional selective scanning captures long-range spatial dependencies with linear computational complexity. An axial- and dilation-aware visual state-space (ADVSS) module is further introduced to adapt the Mamba branch to the directional and periodic characteristics of high-frequency vertical fringes. Meanwhile, the Swin Transformer blocks at the bottleneck and decoder stages perform hierarchical local feature aggregation and facilitate fine-detail recovery during depth decoding. The network is optimized using a compound objective that combines the mean squared error (MSE) and structural similarity index measure (SSIM) losses. Sections 2.1 – 2.4 describe the network architecture and explain the motivation for each component, while Section 2.5 presents the training objective.

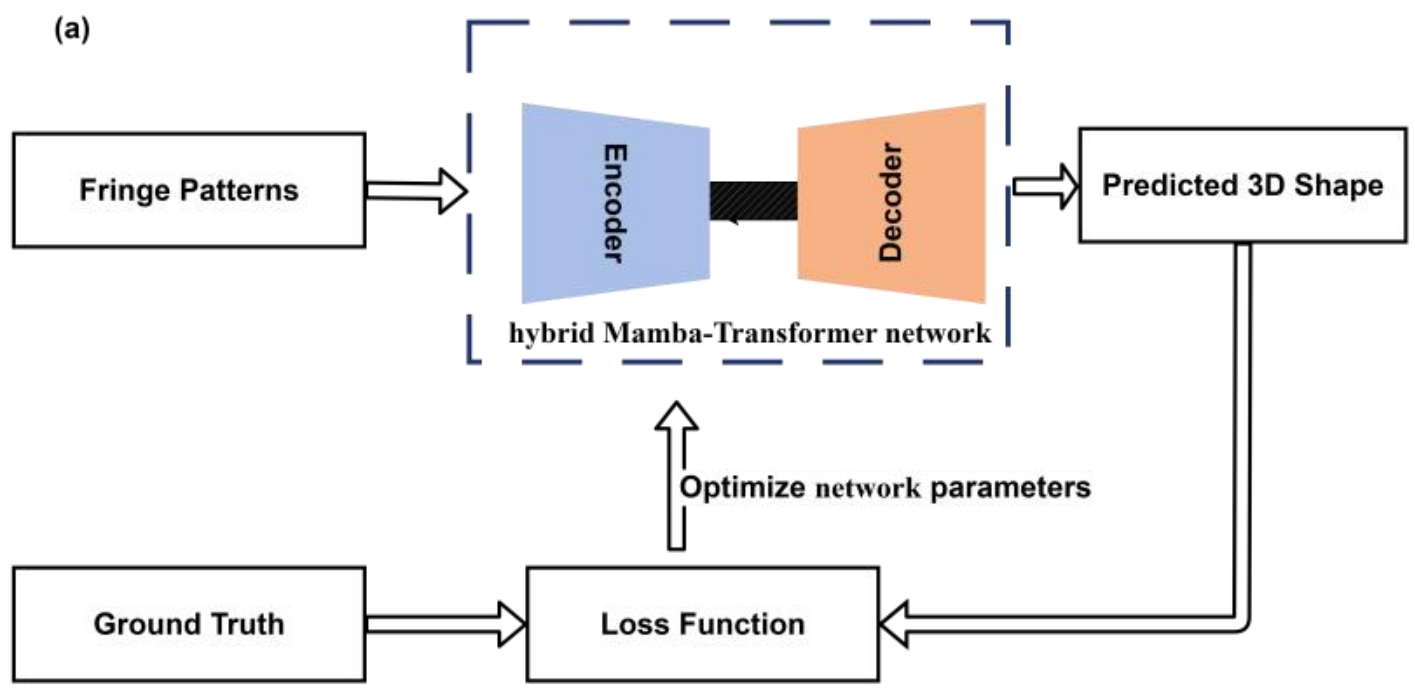

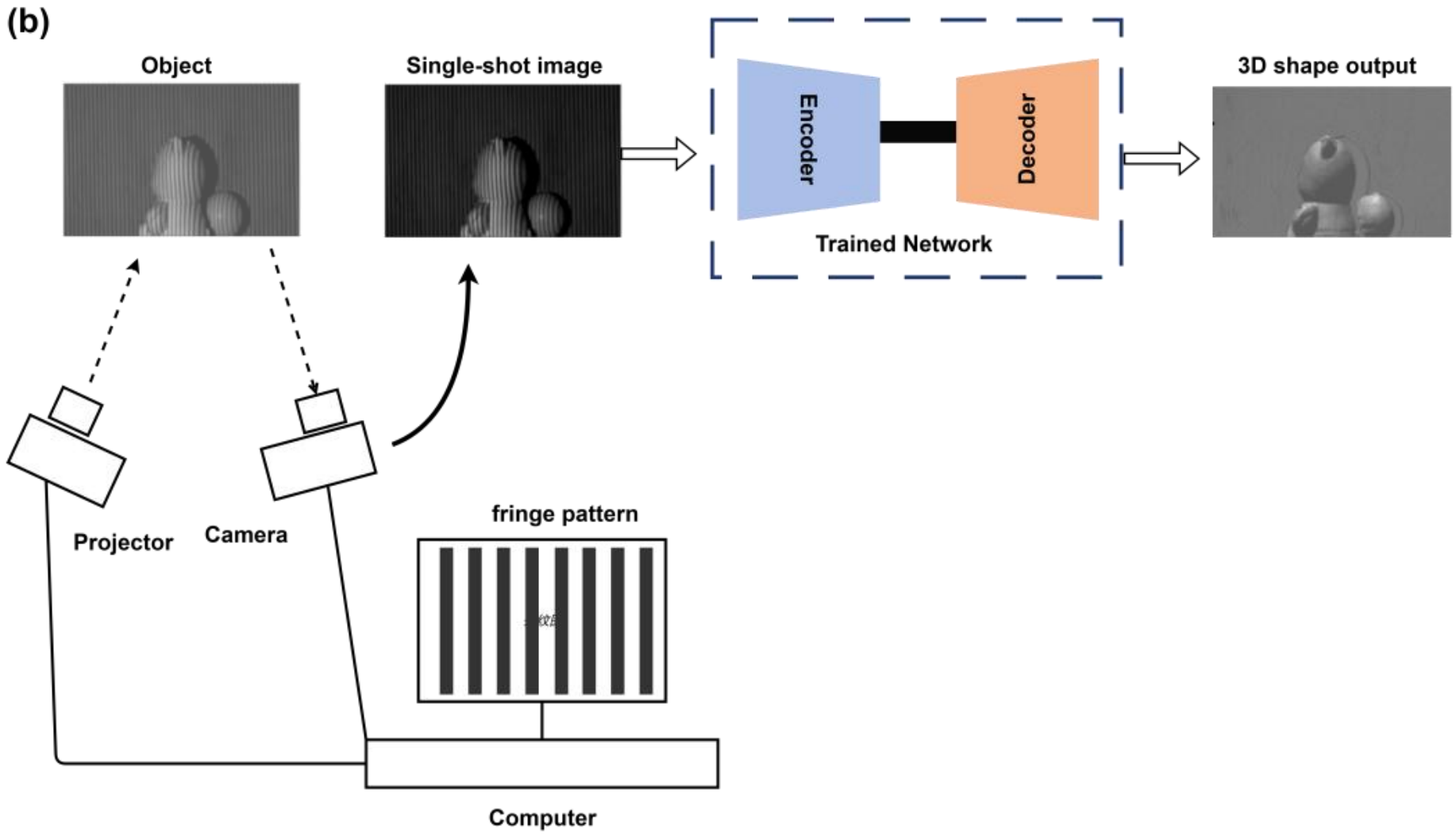

Fig. 1 Proposed method (a) training procedure (b) the reconstruction workflow for a tested object.

## 2.1 Overall network architecture

A hybrid Mamba–Transformer network is developed for three-dimensional depth-map reconstruction from fringe patterns, as illustrated in Fig. 2. The network adopts a hierarchical U-shaped encoder–decoder architecture with attention-gated skip connections to retain informative spatial features during reconstruction.

Given an input fringe tensor $X \subset R^{B\times C_{in}\times H\times W}$ ,where $C_{in}$ depends on the selected input configuration, the patch-embedding layer divides X into non-overlapping 4×4 patches. Each patch is projected onto a C-dimensional feature space, with C=128 used by default. This operation produces an embedded feature map $x^{'} \subset R^{B\times\frac{H}{4}\times\frac{W}{4}\times C}$ .Layer normalization is subsequently applied before the embedded features are passed to the encoder.

The encoder comprises two hierarchical stages, each followed by a patch-merging operation that halves the spatial dimensions and doubles the channel dimension. The first stage, denoted by $L_0$, contains two Swin Transformer blocks with C channels. Alternating window-based multi-head self-attention (W-MSA) and shifted-window multi-head self-attention (SW-MSA) enables these blocks to extract local and cross-window features. The second stage, denoted by $L_1$, comprises four ADVSS blocks with 2C channels. Their four-directional selective scanning captures long-range spatial dependencies with linear complexity with respect to the number of spatial locations, whereas the axial and dilated depth-wise convolutions enhance the representation of directional fringe structures at multiple spatial scales. Following the second patch-merging operation, six Swin Transformer blocks form the bottleneck, processing features at a resolution of H/16×W/16 with 4C channels without further down-sampling.

The decoder progressively restores the spatial resolution through two patch-expansion stages. The first stage up-samples the bottleneck features to H/8×W/8 with 2C channels and processes them using four Swin-Transformer blocks. The second stage further restores the features to H/4×W/4 with C channels and applies two Swin-Transformer blocks. At each stage, the corresponding encoder features are adaptively refined by an attention gate before being concatenated with the upsampled decoder features. The concatenated features are linearly projected and then processed by the Swin-Transformer blocks. This mechanism attenuates less relevant skip-path responses while retaining spatial and geometric information useful for depth reconstruction.

Finally, a 4× final patch-expansion layer restores the feature map to the original spatial resolution of H×W.A $1\times1$ convolution maps the resulting features to a single-channel output ,followed by a sigmoid activation produces a single-channel depth map $\hat{D}\subset R^{B\times1\times H\times W}$ with values normalized to [0,1].

The principal components of the proposed network are described in the following subsections.

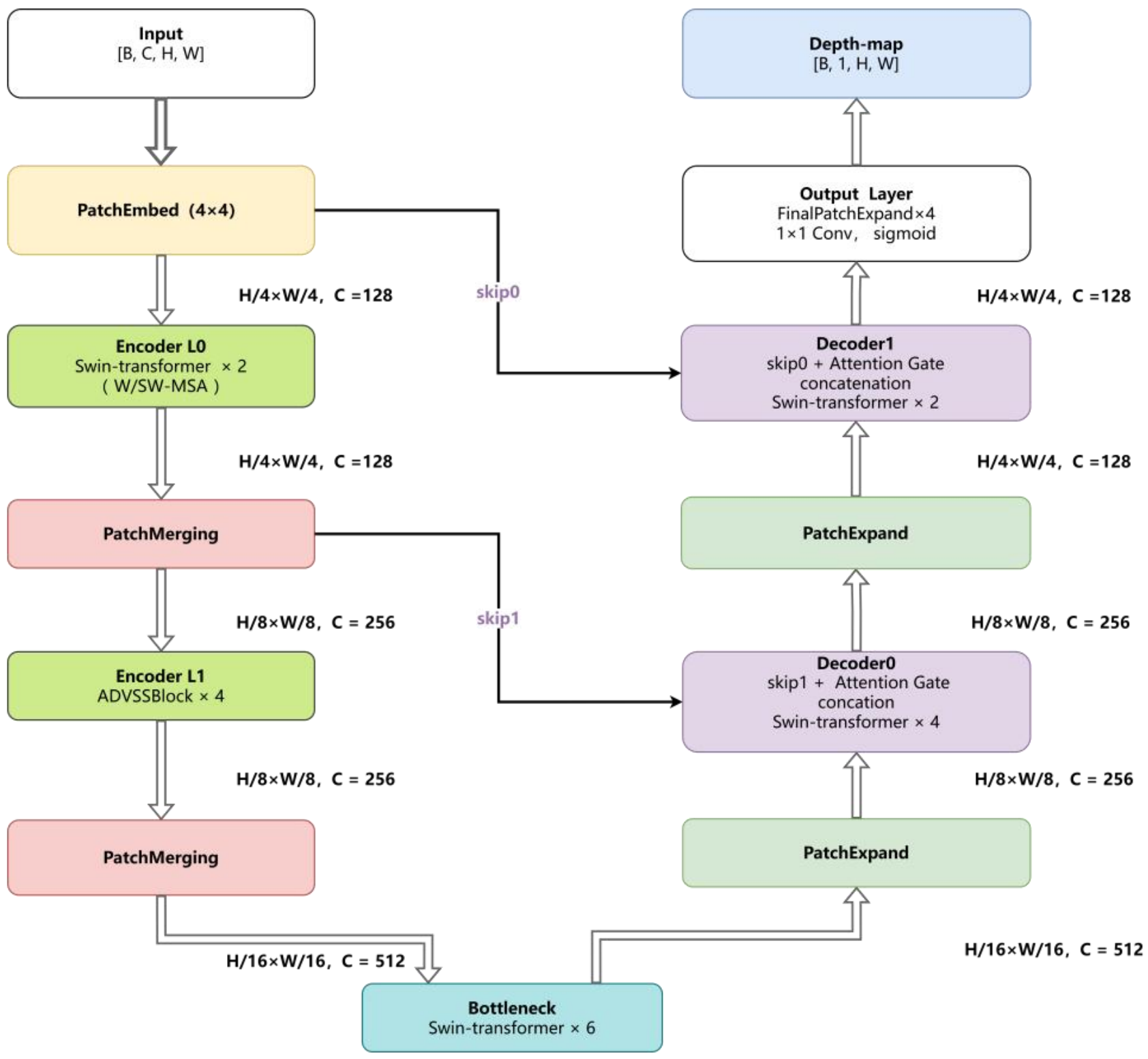

Fig. 2. Overall architecture of the proposed Hybrid Mamba-Transformer network

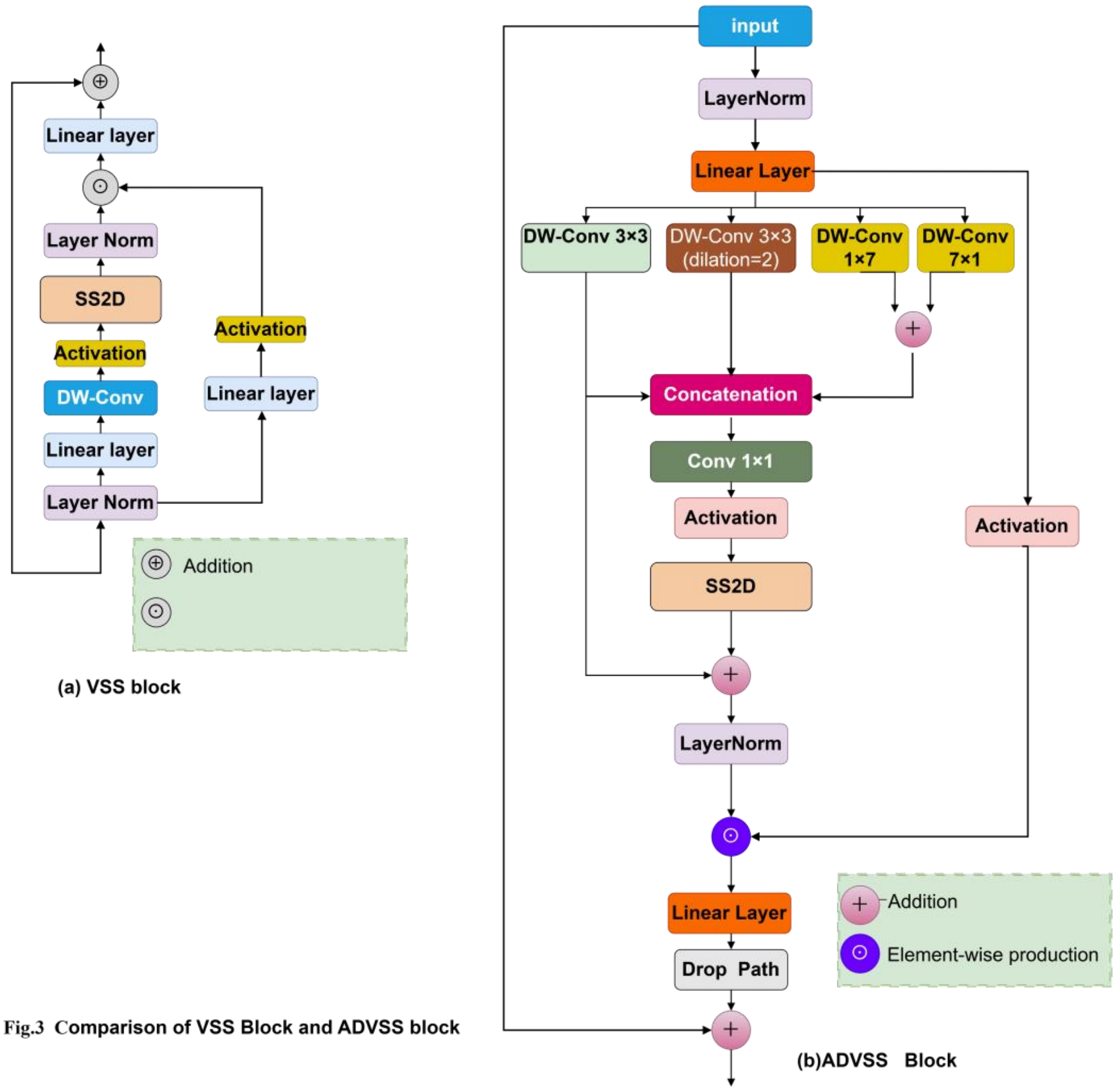

Fig.3 Comparison of VSS Block and ADVSS block

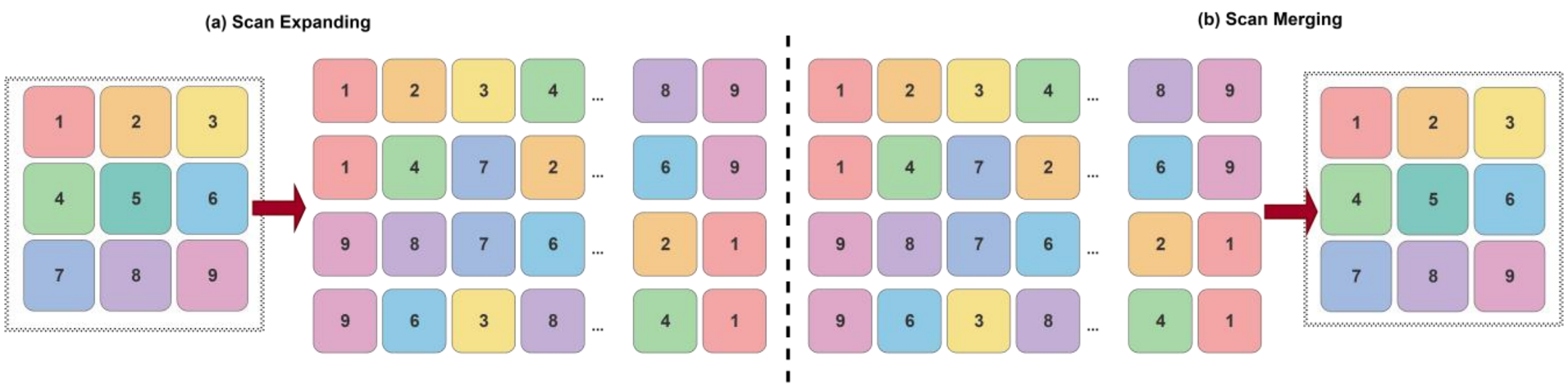

**Fig. 4. Illustration of the SS2D operations: (a) scan expansion and (b) scan merging**

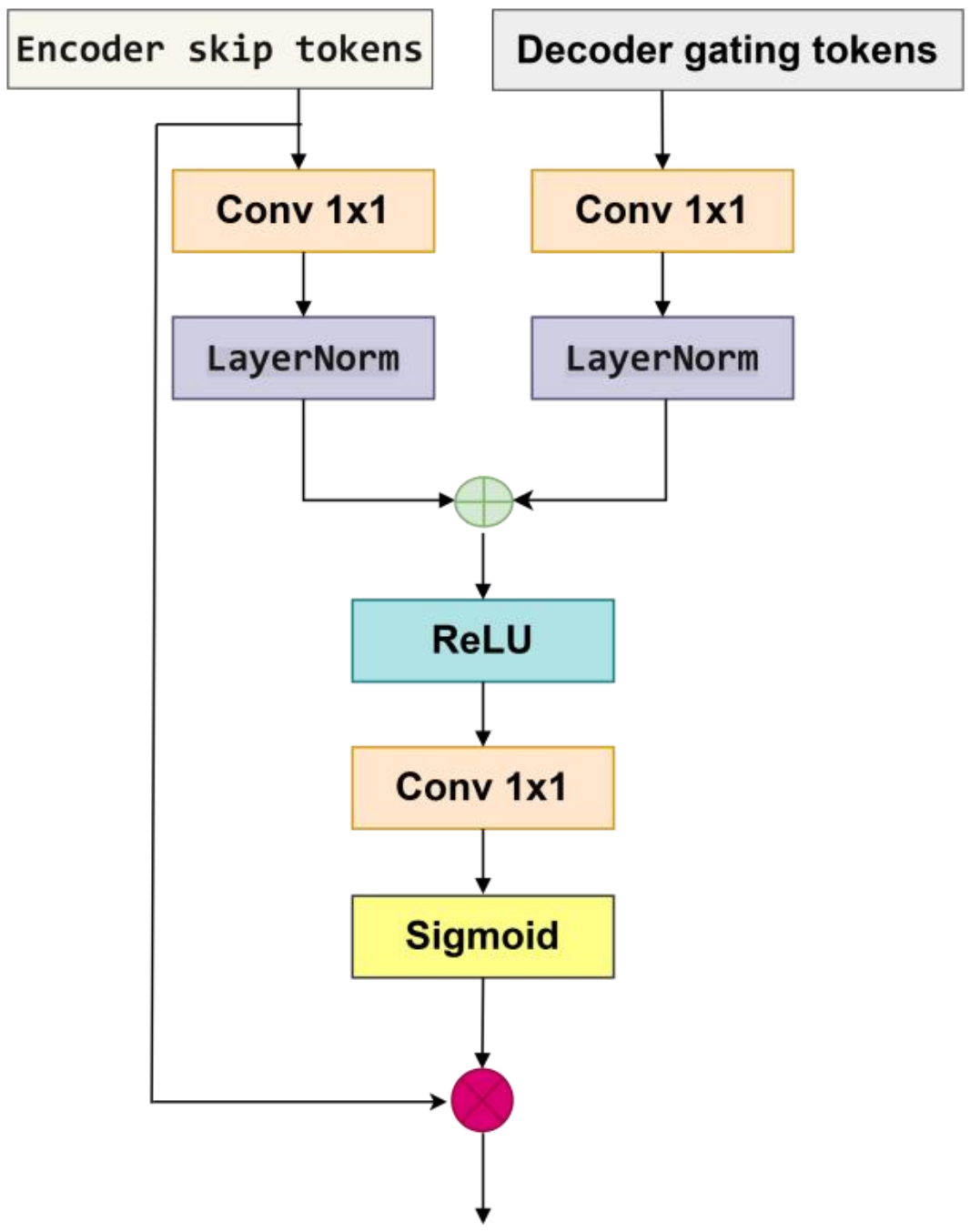

**Fig.5 Attention Gate**

## 2.2 Details of Swin-Transformer Block

Although global self-attention enables Transformers to model long-range dependencies, its computational complexity increases quadratically with the number of input tokens. Its direct application to high-resolution optical images therefore entails substantial memory consumption and computational cost. The Swin Transformer[13] addresses this limitation by alternately employing window-based multi-head self-attention (W-MSA) and shifted-window multi-head self-attention (SW-MSA). W-MSA restricts self-attention to non-overlapping local windows, whereas SW-MSA shifts the window partition between consecutive blocks to facilitate information exchange across neighboring windows. With a fixed window size, this hierarchical attention mechanism achieves approximately linear computational complexity with respect to the number of image patches. It consequently provides an effective balance between local-detail extraction, cross-window feature interaction, and computational efficiency.

Multi-head self-attention (MSA) computes attention weights among input tokens to capture contextual dependencies. Given an input feature matrix $X \subset R^{N \times d}$, where N is the input sequence length and d is the feature dimension, the input is first projected into query, key, and value matrices Q, K, and V as follows:

$$Q = XW^{Q}, K = XW^{K}, V = XW^{V} \tag{1}$$

Here, $W^{Q}, W^{K}, W^{V} \subset R^{d \times d_k}$ are learnable weight matrices. The multi-head self-attention is computed as:

$$Attention(Q, K, V) = soft\max\left(\frac{QK^{T}}{\sqrt{d_k}} + B\right)V \tag{2}$$

where B denotes the relative position bias. In the multi-head setting, several attention heads are computed in parallel and concatenated:

$$MSA(X) = Concat(head_1, \cdots, head_h)W_o \quad (3)$$

With $head_i = Attention(QW_Q^{(i)}, KW_K^{(i)}, VW_V^{(i)})$ and $W_o$ a learnable output projection.

In W-MSA, self-attention is restricted to a fixed-size local window, which substantially reduces computational cost while preserving local structural cues. In SW-MSA, the window partition is shifted relative to that of the preceding layer, thereby enabling interactions across neighboring windows and alleviating the isolation of local regions. A complete Swin Transformer block further incorporates Layer Normalization (LN) and a multilayer perceptron (MLP) with residual connections. Two consecutive blocks can be expressed as

$$\hat{z}^l = \mathrm{W-MSA}\left(LN(z^{l-1})\right) + z^{l-1} \quad (4)$$

$$z^l = \mathrm{MLP}\left(LN(\hat{z}^l)\right) + \hat{z}^l \quad (5)$$

$$\hat{z}^{l+1} = \mathrm{SW-MSA}\left(\mathrm{LN}(z^l)\right) + z^l \quad (6)$$

$$\mathrm{z}^{l+1} = \mathrm{MLP}\left(\mathrm{LN}(\hat{\mathrm{z}}^{l+1})\right) + \hat{\mathrm{z}}^{l+1} \quad (7)$$

By alternating W-MSA and SW-MSA layers, the network can progressively aggregate both local and cross-window representations at multiple scales.

### 2.3 Details of ADVSS block

Although alternating W-MSA and SW-MSA facilitates information exchange across neighboring windows, the effective receptive field of a Swin Transformer expands progressively with network depth. This mechanism may limit the direct modeling of long-range spatial dependencies in high-resolution fringe patterns. To overcome this limitation, an axial–dilated vision state-space (ADVSS) block is introduced. The ADVSS block preserves the linear-complexity selective scanning of the standard vision state-space (VSS) block[19] while enhancing its pre-scan representation through multi-scale and direction-aware depth-wise convolutions.

As shown in Fig. 3, the ADVSS block is derived from the standard VSS block in Fig. 3(a). Its SS2D operator, gating branch, and outer residual connection are retained. The original depth-wise convolution is replaced by three parallel branches, as illustrated in Fig. 3(b): a 3×3 depth-wise convolution for local feature extraction, a dilated 3×3 depth-wise convolution with a dilation rate of 2 for enlarged-context modeling, and a pair of axial 1×7 and 7×1 depth-wise convolutions for capturing horizontally and vertically extended fringe structures. The resulting features are concatenated and fused using a 1×1 convolution before being processed by SS2D. In addition, the local 3×3 convolution response is injected into the scan output before normalization and gating, thereby preserving fine-scale fringe details during global feature aggregation.

As illustrated in Fig. 3(b), the ADVSS block follows a residual formulation. Given an input feature *X*, the block output is

$$Y = X + DropPath\left(F\left(LN(X)\right)\right) \quad (8)$$

where LN( · ) denotes Layer Normalization,DropPath denotes stochastic depth, which randomly drops entire residual branches during training to reduce overfitting and improve model generalization,F(.) denotes the axial–dilated selective-scan operator. Inside F, a linear projection first expands the channel dimension by a factor expand=2 (default) and splits the features into an x-branch and a gating branch z:

$$[x, z] = split(Linear(LN(x))), x, z \in R^{B\times H\times W\times d_{inner}} \tag{9}$$

on the x-branch, three parallel depth-wise convolutions extract multi-scale spatial cues before selective scanning:

$$x_{local} = DWConv_{3\times3}(x), \tag{10}$$

$$x_{mid} = DWConv_{3\times3,d=2}(x), \tag{11}$$

$$x_{axial} = DWConv_{1\times7}(x) + DWConv_{7\times1}(x), \tag{12}$$

Here, the standard 3×3 kernel captures local carrier and edge responses; the dilated 3×3 kernel enlarges the receptive field to neighboring fringe periods,where *d=2* denotes the dilation rate; and the axial 1×7/7×1 convolution kernels emphasize horizontal and vertical mid-range continuity along and across fringe stripes. The three branches are concatenated and fused by a 1×1 convolution followed by a SiLU activation:

$$x_{fused} = SiLU(Conv_{1\times1}(Concat[x_{local}, x_{mid}, x_{axial}])) \tag{13}$$

The fused feature is subsequently fed into the two-dimensional selective scan (SS2D) module for global contextual aggregation. SS2D comprises three successive stages: scan expansion, selective state-space processing using the S6 block, and scan merging. As illustrated in Fig. 4(a), scan expansion unfolds the two-dimensional feature map into four one-dimensional sequences through row-wise and column-wise traversal in both the forward and reverse directions. These complementary sequences are processed in parallel by the S6 block, enabling content-dependent aggregation of long-range spatial information. As shown in Fig. 4(b), the four directional responses are restored to their original orientations, reshaped to the initial spatial layout, and summed to obtain the aggregated feature map. By integrating bidirectional horizontal and vertical dependencies, SS2D provides each spatial position with a global receptive field while maintaining linear computational complexity with respect to the number of spatial locations.

The core of each directional scan is the S6 block, which implements a discrete selective state-space model (SSM). Given a sequence $x \in R^{B\times L\times D}$, the input-dependent parameters Δ, B, and C are obtained by linear projection $linear(x)$,and the continuous parameters are discretized as

$$\overline{A} = exp(\Delta A), \overline{B} = (\Delta A)^{-1}(\exp(\Delta A) - I) \bullet \Delta B \tag{14}$$

The hidden state and output at step t are updated by

$$h_t = \overline{A}h_{t-1} + \overline{B}x_t, y_t = Ch_t + Dx_t \quad (15$$

and the full sequence is obtained by concatenating $\{y_t\}_{t=1}^{L}$ , Through input-dependent parameterization and discretization, the S6 block selectively retains, propagates, or attenuates information along long sequences, thereby capturing long-range dependencies beyond the fixed local windows of window-based self-attention.

After four-directional scanning and merging, the local branch is added back before normalization and gating:

$$y = SS2D(x_{fused}) + x_{local} \tag{16}$$

$$y = LN(y) \odot SiLU(z) \tag{17}$$

$$\bar{Y} = linear(y) \tag{18}$$

where ⊙ denotes element-wise multiplication. This residual injection allows fine-scale local features to reach the gating stage directly, while the z-branch adaptively attenuates uninformative responses. No additional MLP is introduced after selective scanning; channel interactions are instead modeled by the input and output projections and the 1×1 fusion layer.

Overall, the ADVSS block integrates axial–dilated local encoding with four-directional selective scanning, efficiently capturing long-range dependencies while preserving fringe-sensitive spatial details for subsequent surface-height regression.

### 2.4 Details of Attention Gate

Skip connections in encoder–decoder architectures recover spatial information lost during downsampling. However, directly concatenating encoder and decoder features may introduce redundant or task-irrelevant information. An attention gate (AG) is therefore applied to adaptively refine the encoder features before feature fusion, as illustrated in Fig. 5.

The AG receives the encoder skip tokens and the decoder gating tokens. Both inputs are projected into a common intermediate dimension using 1×1 convolutions followed by layer normalization.The resulting feature maps are then combined by element-wise addition and passed through a ReLU activation. A subsequent 1×1 convolution , followed by a sigmoid function, produces a spatial attention map with coefficients in [0,1]. Finally, this attention map is applied to the encoder skip tokens via element-wise multiplication, yielding the gated skip features.Guided by the decoder representation, the AG attenuates less relevant encoder responses while retaining informative spatial features, thereby enabling more selective multi-scale fusion and facilitating the recovery of fine geometric details.

### 2.5 Compound Loss Function

The network aims to learn a nonlinear mapping from an input fringe pattern to the corresponding three-dimensional surface-height map by optimizing its trainable parameters. To simultaneously enforce point-wise accuracy and local structural consistency, a compound loss combining mean squared error (MSE) and structural similarity (SSIM) is adopted:

$$\mathcal{L} = \alpha\mathcal{L}_{\text{MSE}} + \beta\mathcal{L}_{\text{SSIM}} = \alpha\frac{1}{N}\sum_{i=1}^{N}\left(\widehat{D}_i - D_i\right)^2 + \beta\left[1 - \text{SSIM}\left(\widehat{\mathbf{D}}, \mathbf{D}\right)\right], \tag{19}$$

Where $\widehat{\mathbf{D}}$ and $\mathbf{D}$ denote the predicted and ground-truth height maps, respectively, and N is the number of pixels. The weighting coefficients were set to α =0.80.and β=0.20 based on sensitivity experiments.

The MSE term penalizes point-wise height deviations and therefore promotes accurate absolute reconstruction. However, because errors are squared independently at each pixel, MSE is relatively sensitive to outliers and does not explicitly characterize local structural consistency. The SSIM loss complements MSE by encouraging the preservation of local surface structures and geometric variations. SSIM is computed with a data range of 1.0, consistent with the normalized height maps used during training. Consequently, the compound objective improves point-wise reconstruction accuracy without sacrificing the structural fidelity of the recovered surface.

To calculate the SSIM between two fringe-pattern images, we can use the following formula[10]:

$$S(\mathbf{x}, \mathbf{y}) = f(l(\mathbf{x}, \mathbf{y}), c(\mathbf{x}, \mathbf{y}), s(\mathbf{x}, \mathbf{y})) \tag{20}$$

Where brightness *l(x,y)*, contrast *c(x,y)* and structure *s(x,y)* are respectively calculated using the following formulas:

$$l(x, y) = \frac{2\mu_x\mu_y}{\mu_x^2 + \mu_y^2} \tag{21}$$

$$c(x, y) = \frac{2\sigma_x\sigma_y}{\sigma_x^2 + \sigma_y^2} \tag{22}$$

$$s(x, y) = \frac{\sigma_{xy}}{\sigma_x\sigma_y} \tag{23}$$

The SSIM can be expressed as follow:

$$\mathrm{SSIM}\ (x, y) = \frac{(2\mu_x\mu_y + c_1)(2\sigma_{xy} + c_2)}{(\mu_x^2 + \mu_y^2 + c_1)(\sigma_x^2 + \sigma_y^2 + c_2)} \tag{24}$$

Where, $\mu_x$, $\mu_y$ are the average values of *x* and *y* pattern image respectively, $\sigma_x^2$, $\sigma_y^2$ is the variance of the two images, and $\sigma_{xy}$ is the covariance of *x* and *y*. $c_1 = (k_1L)^2$ and $c_2 = (k_2L)^2$, which are constants used to maintain stability. *L* is the dynamic range of pixel values, where $k_1$=0.01 and $k_2$=0.03. They are adopted as the conventional default values to prevent numerical instability when the local mean or variance approaches zero; these constants are not re-tuned on the present dataset.The value of SSIM is limited between 0 and 1. In this study, *x* and *y* correspond to the predicted and ground-truth normalized depth maps, respectively.

## 3. Dataset and Experiments

In this section, we present a series of experiments to validate the effectiveness of our method. Firstly, in Section 3.1, we provide a concise overview of the dataset used for network training, as well as the details of the training process. Next, in Section 3.2, we conduct ablation experiments to investigate the impact of different network structures and loss functions on the training outcomes. Furthermore, in Section 3.3, we compare our method with hNet, SwinConvUNet, UHRNet, and HiDNet, utilizing a dataset obtained from Ref. [8]. Finally, in Section 3.4, we perform 3D reconstruction tests on objects that are not included in the training set. This allows us to comprehensively demonstrate the effectiveness of our proposed method.

### 3.1 Dataset and training

The dataset reported in Ref. [8] was used to train and evaluate the proposed method. It contains fringe images of objects with diverse geometries, dimensions, and surface characteristics, providing representative samples for fringe-to-depth reconstruction. The images were acquired using a structured-light projection system at a spatial resolution of 768×512 pixels. The original dataset comprises 3,070 samples, with each sample containing 60 images acquired by the left and right cameras—30 images per camera—and the corresponding depth labels. Only the data acquired by the left camera were used in this study, resulting in 3,070 fringe–depth pairs.

For each sample, the ninth frame, corresponding to a high-frequency vertical sinusoidal pattern containing 128 fringes, was selected as the network input. This frame is the first pattern in the eight-step phase-shifting sequence. The 29th frame was used as the reference pattern. The ground-truth depth maps were generated from multiple images within each acquisition sequence using a rational-fraction-based reconstruction method without an intermediate matching procedure. Representative samples are presented in Fig. 6, where the first and second columns show the selected input fringe patterns and their corresponding ground-truth depth maps, respectively.

Of the 3,070 samples, a fixed subset of 74 samples (2.4%) was reserved as the independent test set and stored in fringe_test.h5 and Z_test.h5. The remaining 2,996 samples were stored in fringe.h5 and Z.h5 and randomly divided into 2,697 training samples (87.9% of the complete dataset) and 299 validation samples (9.7%). The test samples were excluded from both network optimization and model selection.

The training data were stored as chunked and compressed HDF5 arrays rather than PyTorch .pt files. During dataset initialization, only metadata, including the array dimensions and number of samples, was read; individual fringe images and depth maps were subsequently loaded on demand by index. Each DataLoader worker independently opened the corresponding HDF5 files, preventing complete float32 arrays from being duplicated across worker processes. Compared with loading the entire dataset into memory, which would require approximately 9 GB, this sample-wise loading strategy substantially reduced memory consumption and supported efficient parallel data loading.

During training, the network parameters were optimized through back-propagation using the training set. Performance on the validation set was monitored throughout training, and the model yielding the lowest validation loss was retained for subsequent evaluation. The selected model was then evaluated on the independent test set. Because the test samples were not involved in either parameter optimization or checkpoint selection, the resulting performance provides an independent assessment of the generalization capability of the proposed method.

All experiments were performed on a workstation equipped with an Intel Xeon Platinum 8375C processor (2.90 GHz), 251 GB of RAM, and an NVIDIA GeForce RTX 4090 graphics card with 24 GB of memory.The software environment comprised Ubuntu 24.04.2, Python 3.10.21 The training code was implemented in PyTorch 2.0.1 with CUDA 11.8 and optimized with the Adam optimizer. The batch size was set to 4, and each network was trained for 400 epochs in full precision without early stopping. The learning rate was initialized to $1\times10^{-4}$ and annealed by a cosine schedule over 400 epochs to a minimum of $6\times10^{-5}$ . The loss combined the Mean Squared Error (MSE) and the Structural Similarity Index Measure (SSIM), with weights of 0.8 and 0.2, respectively. Of the samples used for model fitting, 10% were randomly held out as a validation set to monitor training.

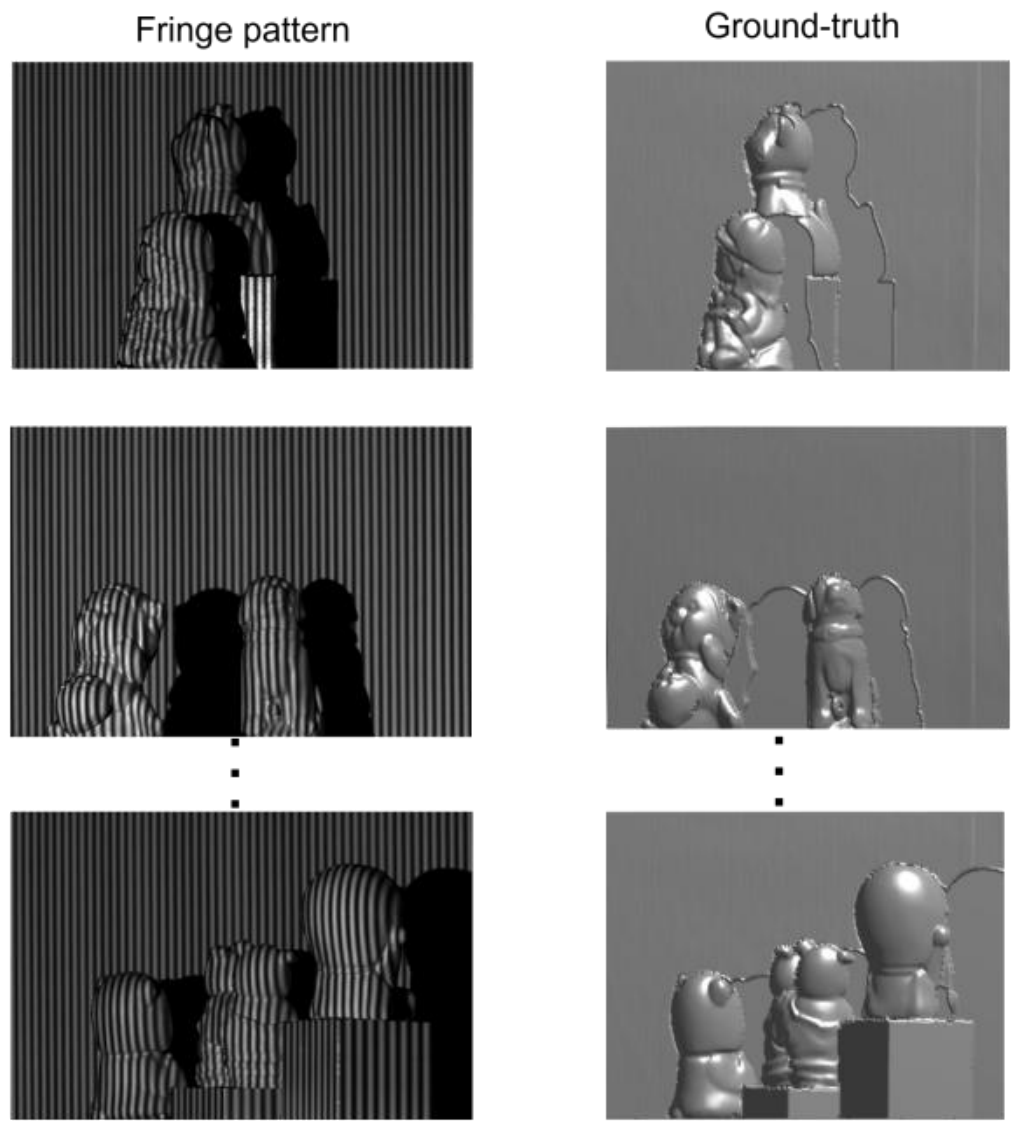

Fig. 6 Partial samples used in datasets
Left column: fringe-patterns Right column: ground-truth

### 3.2 Evaluation metrics

We employ four metrics to quantitatively assess reconstruction quality[7-8]. The Mean Absolute Error (MAE) and Root Mean Square Error (RMSE) quantify physical deviation in millimeters

The SSIM can be calculated using Eq.24 ,MAE and RMSE can be calculated using the following formula:

$$\mathrm{MAE} = \frac{1}{\mathrm{N}}\sum_{i=1}^{N}\left|y_i - y_i^p\right| \tag{25}$$

$$RMSE = \sqrt{\frac{1}{N}\sum_{i=1}^{N}\left(y_i - y_i^p\right)^2} \tag{26}$$

Where, $y_i$ is the ground-truth depth, $y_i^p$ is the predicted depth at pixel *i*, and *N* is the effective number of pixel points(except background pixels).

The Absolute Relative Error (AbsRel) measures error proportional to depth:

$$AbsRel = \frac{1}{N}\sum_{i=1}^{N}\frac{\left|y_i^p - y_i\right|}{y_i} \tag{27}$$

### 3.3 Ablation experiment

A series of ablation experiments was conducted to investigate how the principal architectural components affect single-shot 3D reconstruction and to determine the most effective configuration. All variants were trained and evaluated on the same fringe-pattern dataset using an identical experimental protocol and evaluation criteria.

With reference to the hybrid Mamba–Transformer architecture shown in Fig. 2, the decoder configuration was kept unchanged. The ablations focused on the encoder stages ($L_0$ and $L_1$), the bottleneck ($L_2$), and the skip-connection strategy. The evaluated configurations are described below.

**Architecture ablations**

**Mode A:** $L_0$, $L_1$, and $L_2$ contain 2, 4, and 6 Swin Transformer blocks, respectively. Attention-gated skip connections are employed.

**Mode B:** $L_0$ and $L_2$ contain 2 and 6 Swin Transformer blocks, respectively, whereas $L_1$ contains 4 VSS blocks. Attention-gated skip connections are retained.

**Mode C:** $L_0$ and $L_1$ contain 2 and 4 VSS blocks, respectively, whereas $L_2$ contains 6 Swin Transformer blocks. Attention-gated skip connections are retained.

**Mode D:** $L_0$ and $L_1$ contain 2 and 4 Swin Transformer blocks, respectively, whereas $L_2$ contains 6 VSS blocks. Attention-gated skip connections are retained.

**Mode E:** The VSS blocks in $L_1$ of Mode B are replaced with 4 proposed ADVSS blocks. $L_0$ and L2 retain 2 and 6 Swin Transformer blocks, respectively, and attention-gated skip connections are used. This configuration represents the complete proposed model.

**Mode F:** Mode F has the same encoder and bottleneck configuration as Mode E, but the attention gates are removed. The encoder features are therefore concatenated directly with the corresponding decoder features.

**Component ablations of the ADVSS block**

The following variants were derived from Mode E to evaluate the individual components of the ADVSS block.

**Mode G:** The axial-convolution branch is removed from each ADVSS block in $L_1$.

**Mode H:** The dilated-convolution branch is removed from each ADVSS block in $L_1$.

**Mode I:** The post-scan local residual injection is removed from each ADVSS block in $L_1$, while the local 3×3 branch used for pre-scan feature fusion is retained.

All other architectural and training settings in Modes G–I are identical to those of Mode E. Modes A–I use a single deformed fringe pattern as the network input.

**Input-channel ablations**

To examine the effect of incorporating additional input information, the five-channel input configuration corresponding to Method 4 of Dinh et al. [7] was further evaluated. It comprises the deformed fringe pattern, reference fringe pattern, normalized coordinate maps U and V, and difference map.

$$\Delta I = \frac{I - I_{ref}}{2} \tag{28}$$

Because of the additional computational cost, this protocol was applied only to Modes A, B, and E, producing Modes J, K, and L, respectively.

Each mode was evaluated on the same test set by comparing the predicted height maps with their corresponding ground-truth maps. Reconstruction performance was quantified using mean absolute error (MAE), root-mean-square error (RMSE), absolute relative error (AbsRel), and structural similarity index measure (SSIM). Lower MAE, RMSE, and AbsRel values indicate greater reconstruction accuracy, whereas a higher SSIM indicates better structural agreement.

All metrics were calculated over a common set of valid pixels defined by morphologically eroded ground-truth masks, with a mean valid-pixel coverage of 90.6%±5.4%. The preferred direction of each metric is indicated by an arrow, and the best result is highlighted in bold.

Table 2 summarizes the architecture ablations.

**Table 2. Architecture ablations of the encoder, bottleneck, and skip connection on the test set.**

| Model | channels | Parameters(M) | MAE(mm)↓ | RMSE(mm)↓ | AbsRel↓ | SSIM↑ |
|---|---|---|---|---|---|---|
| A | **1** | 27.89 | 0.369 | 1.089 | 0.093 | 0.9920 |
| B | 1 | 27.00 | 0.215 | 0.643 | 0.051 | 0.9973 |
| C | 1 | 26.94 | 0.235 | 0.679 | 0.059 | 0.9969 |
| D | 1 | 21.28 | 0.199 | 0.593 | 0.059 | 0.9966 |
| E | 1 | **30.20** | **0.195** | **0.548** | **0.043** | **0.9985** |
| F | 1 | 30.12 | 0.236 | 0.613 | 0.062 | 0.9967 |

Compared with the all-Swin baseline (Mode A), replacing the Swin blocks at the L1 stage with VSS blocks (Mode B) reduces the MAE from 0.369 to 0.215 mm and the RMSE from 1.089 to 0.643 mm. The AbsRel decreases from 0.093 to 0.051, while the SSIM increases from 0.9920 to 0.9973. Extending the VSS blocks to both the L0 and L1 stages (Mode C) does not yield further improvements. This result suggests that selective scanning is more effective at the intermediate-resolution L1 stage than at the highest-resolution L0 stage.

Introducing VSS blocks only at the bottleneck (Mode D) produces lower MAE and RMSE values than Mode B, but results in higher AbsRel and lower SSIM. Mode E achieves the best overall performance, with an MAE of 0.195 mm, an RMSE of 0.548 mm, an AbsRel of 0.043, and an SSIM of 0.9985. Compared with Mode B, these results indicate that incorporating axial, dilated, and local feature encoding into the L1 VSS blocks improves both geometric accuracy and structural fidelity.

Furthermore, Mode E consistently outperforms Mode F, reducing the MAE from 0.236 to 0.195 mm, the RMSE from 0.613 to 0.548 mm, and the AbsRel from 0.062 to 0.043, while increasing the SSIM from 0.9967 to 0.9985. The attention gates are the only architectural difference between these two Modes, this comparison confirms their effectiveness in suppressing irrelevant skip-connection features and improving depth reconstruction accuracy.

Table 3 presents component ablation of the ADVSS block.

**Table 3. Component ablations of the ADVSSBlock (relative to Mode E)**

| Model | channels | Parameters(M) | MAE(mm)↓ | RMSE(mm)↓ | AbsRel↓ | SSIM↑ |
|---|---|---|---|---|---|---|
| E | 1 | 30.20 | 0.195 | 0.548 | 0.043 | 0.9985 |
| G | 1 | 29.12 | 0.205 | 0.580 | 0.044 | 0.9978 |
| H | 1 | 29.13 | 0.199 | 0.5489 | 0.055 | 0.9971 |
| I | 1 | 30.20 | 0.196 | 0.608 | 0.058 | 0.9971 |

The complete ADVSS block (Mode E) achieves the best performance across all four metrics. Removing the axial-convolution branch (Mode G) increases the MAE from 0.195 to 0.205 mm and the RMSE from 0.548 to 0.580 mm, while reducing the SSIM from 0.9985 to 0.9978. These results suggest that axial convolutions contribute to modeling elongated and directionally continuous fringe structures.

Removing the dilated-convolution branch (Mode H) has little effect on the RMSE, which increases from 0.548 to 0.5489 mm, but increases AbsRel from 0.043 to 0.055 and decreases SSIM to 0.9971. The dilated branch therefore appears to contribute primarily to relative-height consistency and structural preservation. Removing the post-scan local residual injection (Mode I) results in only a marginal increase in MAE, from 0.195 to 0.196 mm, but produces the highest

RMSE and AbsRel among the four variants, at 0.608 mm and 0.058, respectively. This result suggests that the local residual pathway helps limit large reconstruction errors that may not be adequately reflected by MAE alone. Overall, the results support retaining the axial, dilated, and local residual mechanisms in the complete ADVSS block.

Table 4 compares three matched architecture pairs under single- and five-channel input protocols. Modes A, B, and E use a single deformed fringe image, whereas Modes J, K, and L employ the five-channel configuration corresponding to Method 4 of Dinh et al. [7]. The latter comprises the deformed fringe image, reference fringe image, difference map, and two normalized geometric-coordinate maps. Bold values indicate the better result within each matched pair (A–J, B–K, and E–L).

Replacing the single-channel input projection with its five-channel counterpart introduces approximately $8.2\times10^{3}$ additional parameters, corresponding to only 0.008 M parameters. Consequently, the rounded parameter counts differ by no more than 0.01 M within each pair.

**Table 4. Input-channel ablations under matched architectures.**

| Model | Architecture | channels | Parameters(M) | MAE(mm) | RMSE(mm) | AbsRel | SSIM |
|---|---|---|---|---|---|---|---|
| A | All-Swin | 1 | 27.89 | 0.369 | 1.089 | 0.093 | 0.9920 |
| J | All-Swin | 5 | **27.90** | **0.294** | **1.077** | **0.067** | **0.9948** |
| B | L1–VSS | 1 | **27.00** | **0.215** | **0.643** | **0.051** | **0.9973** |
| K | L1 – VSS | 5 | 27.01 | 0.258 | 0.749 | 0.067 | 0.9957 |
| E | L1–ADVSS | 1 | **30.20** | **0.195** | **0.548** | **0.043** | **0.9985** |
| L | L1–ADVSS | 5 | 30.21 | 0.205 | 0.613 | 0.056 | 0.9967 |

The five-channel protocol improves all four metrics for the all-Swin architecture. From Mode A to Mode J, the MAE decreases from 0.369 to 0.294 mm, the RMSE decreases from 1.089 to 1.077 mm, AbsRel decreases from 0.093 to 0.067, and SSIM increases from 0.9920 to 0.9948. In contrast, the same protocol degrades the performance of the two selective-scan architectures. From Mode B to Mode K, the MAE increases from 0.215 to 0.258 mm and the RMSE increases from 0.643 to 0.749 mm. Similarly, from Mode E to Mode L, the MAE increases from 0.195 to 0.205 mm and the RMSE increases from 0.548 to 0.613 mm. AbsRel and SSIM exhibit the same performance trend in both comparisons.

These results suggest an interaction between the input representation and encoder architecture. The all-Swin model benefits from the explicitly supplied reference and geometric information. By contrast, the VSS- and ADVSS-based models already capture long-range dependencies from a single fringe image through selective scanning, and the additional channels do not improve their performance under the present training protocol. Because the parameter increase is negligible, the degradation observed for Modes K and L is unlikely to result from insufficient model capacity. Instead, it may reflect a mismatch between simple channel concatenation and the current selective-scan feature-embedding strategy. Further experiments would be required to establish the underlying mechanism.

The relative ranking of the three architectures remains unchanged under both input protocols. In terms of MAE, $L_1$–ADVSS outperforms $L_1$–VSS, which in turn outperforms the all-Swin configuration: 0.195, 0.215, and 0.369 mm for the single-channel protocol and 0.205, 0.258, and 0.294 mm for the five-channel protocol, respectively. Mode L also outperforms Mode K in RMSE, AbsRel, and SSIM, indicating that the ADVSS refinement remains beneficial with five-channel input. Nevertheless, Mode L remains less accurate than the single-channel Mode E. Mode E is therefore adopted for the principal experiments, while the five-channel configuration of Dinh et al. [7] is retained as an input-protocol ablation.

Overall, the ablation results identify the complete ADVSS block at the intermediate-resolution $L_1$ stage as the most effective configuration among those evaluated, combining local fringe-feature preservation with long-range dependency modeling. Mode E achieves the best overall performance using only one deformed fringe image, without additional reference or coordinate channels. This conclusion is further supported by the performance degradation observed after removing the axial, dilated, or local residual component. Moreover, the additional input channels do not improve the VSS- or ADVSS-based models under the present training protocol, suggesting that selective scanning captures sufficient spatial context from the single-channel input.

**3.4 Quantitative and qualitative comparisons with representative methods**

The proposed method was compared with four representative reconstruction networks: hNet, SwinConvUNet, UHRNet, and HiDNet. All baseline models were re-implemented according to their original architectures, with their input layers adapted where necessary to accept the single-channel fringe images used in this study. Each model takes one deformed fringe pattern as input and predicts a normalized height map, which is subsequently converted to millimetres for quantitative evaluation. All models were trained and tested using the dataset constructed from the publicly available dataset described in Section 3.1

The experiments were conducted on a workstation equipped with an Intel Xeon Platinum 8375C CPU at 2.90 GHz, 251 GB of system memory, and one NVIDIA GeForce RTX 4090 GPU with 24 GB of memory. The software environment comprised Ubuntu 24.04.2, Python 3.10.21, and PyTorch 2.0.1 with CUDA 11.8.Unless otherwise specified, all methods were trained using Adam with an initial learning rate of $1\times10^{-4}$, a batch size of 4, and 400 epochs. The learning rate was adjusted using a cosine annealing schedule and decreased to $6\times10^{-5}$ by the end of training. The remaining experimental settings were kept consistent with those used in the ablation studies.

Figure 7 presents the training and validation RMSE curves. All methods exhibit a rapid reduction in RMSE during the initial training stage and broadly approach convergence after approximately 250 epochs, although occasional fluctuations remain. The proposed method achieves the lowest final RMSE on both the training and validation sets, at 0.236 and 0.607 mm, respectively. Its validation curve is also comparatively stable, particularly during the first 80 epochs, when the baseline methods exhibit several pronounced fluctuations. These results indicate that the proposed architecture provides more accurate and stable optimization under the common training protocol.

A quantitative comparison was further performed by evaluating the reconstructed height maps against their corresponding ground-truth maps on the test set. Four complementary metrics were employed: mean absolute error (MAE), root mean square error (RMSE), absolute relative error (AbsRel), and structural similarity index measure (SSIM). The inference time was also recorded . It was measured on the same NVIDIA GeForce RTX 4090 GPU with a batch size of 1. Before timing, each model was executed for 50 warm-up iterations, followed by 200 timed iterations in evaluation mode with gradient computation disabled. torch.cuda.synchronize() was called immediately before and after each timed forward pass to ensure completion of all asynchronous CUDA operations. The reported inference time is the mean time required to process one fringe image, excluding data loading and result post-processing.The quantitative comparison results are summarized in Table 5.

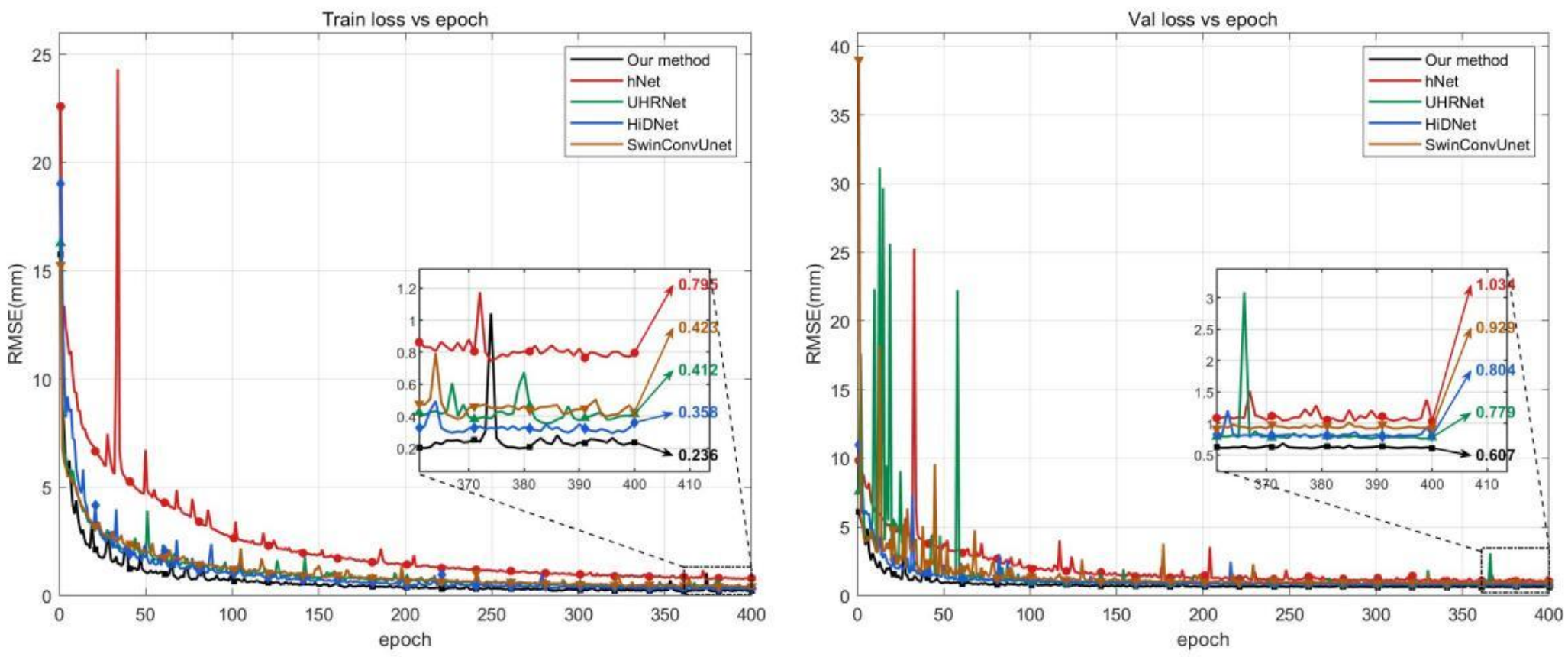

Fig. 7.Training and validation RMSE curves of our method and four representative methods(hNet, SwinConvUNet, UHRNet, and HiDNet). The insets show the final 40 epochs and the corresponding final RMSE values.
Left: training set; right: validation set.
Insets show the last 40 epochs with final RMSE values (mm).

UHRNet has the largest parameter count at 30.363 M, followed closely by the proposed method at 30.225 M, whereas HiDNet is the most compact model at 8.121 M. hNet provides the shortest inference time of 0.0092 s per fringe pattern image, while the proposed method requires 0.0246 s. Nevertheless, lower model complexity or shorter inference time does not necessarily result in greater reconstruction accuracy. Despite its relatively short inference time, hNet produces the highest RMSE and lowest SSIM among the evaluated methods.

The proposed method achieves an MAE of 0.195 mm, an RMSE of 0.548 mm, an AbsRel of 0.043, and an SSIM of 0.9985, outperforming all four baseline methods. Its RMSE is 72.0%, 67.6%, 65.2%, and 59.6% lower than those of hNet, SwinConvUNet, UHRNet, and HiDNet, respectively. Although the proposed method has the longest inference time, it is only 1.5 and 2.2 ms slower per image than SwinConvUNet and UHRNet, respectively. These results demonstrate that the proposed method offers the highest reconstruction accuracy among the evaluated networks, at the cost of a modest increase in inference latency.

**Table5. Quantitative comparison of the proposed method with four representative methods on the test set. The best result for each metric is shown in bold.**

| method | MAE(mm) | RMSE(mm) | AbsRel | SSIM | Number of parameters(M) | Inference time (s) |
|---|---|---|---|---|---|---|
| Proposed method | **0.195** | **0.548** | **0.043** | **0.9985** | 30.225 | 0.0246 |
| hNet | 0.720 | 1.956 | 0.305 | 0.9624 | 8.631 | **0.0092** |
| SwinConvUNet | 0.591 | 1.693 | 0.181 | 0.9672 | 11.022 | 0.0231 |
| UHRNet | 0.571 | 1.574 | 0.179 | 0.9751 | 30.363 | 0.0224 |
| HiDNet | 0.450 | 1.357 | 0.140 | 0.9868 | **8.121** | 0.0212 |

Figure 8 presents a representative single-object reconstruction. The input fringe pattern and ground-truth height map are shown in Figs. 8(a) and (b), respectively, while Figs. 8(c)–(g) show the reconstructions obtained using hNet, SwinConvUNet, UHRNet, HiDNet, and our method. Figure 8(h) compares the absolute-error profiles sampled along the red dotted line in Fig. 8(b). All methods recover the overall shape of the sculpture. However, hNet exhibits the largest deviations, while SwinConvUNet, UHRNet, and HiDNet retain varying levels of localized error. The proposed method produces the lowest error over most of the sampled profile and more faithfully preserves fine surface details, consistent with the test-set results in Table 5.

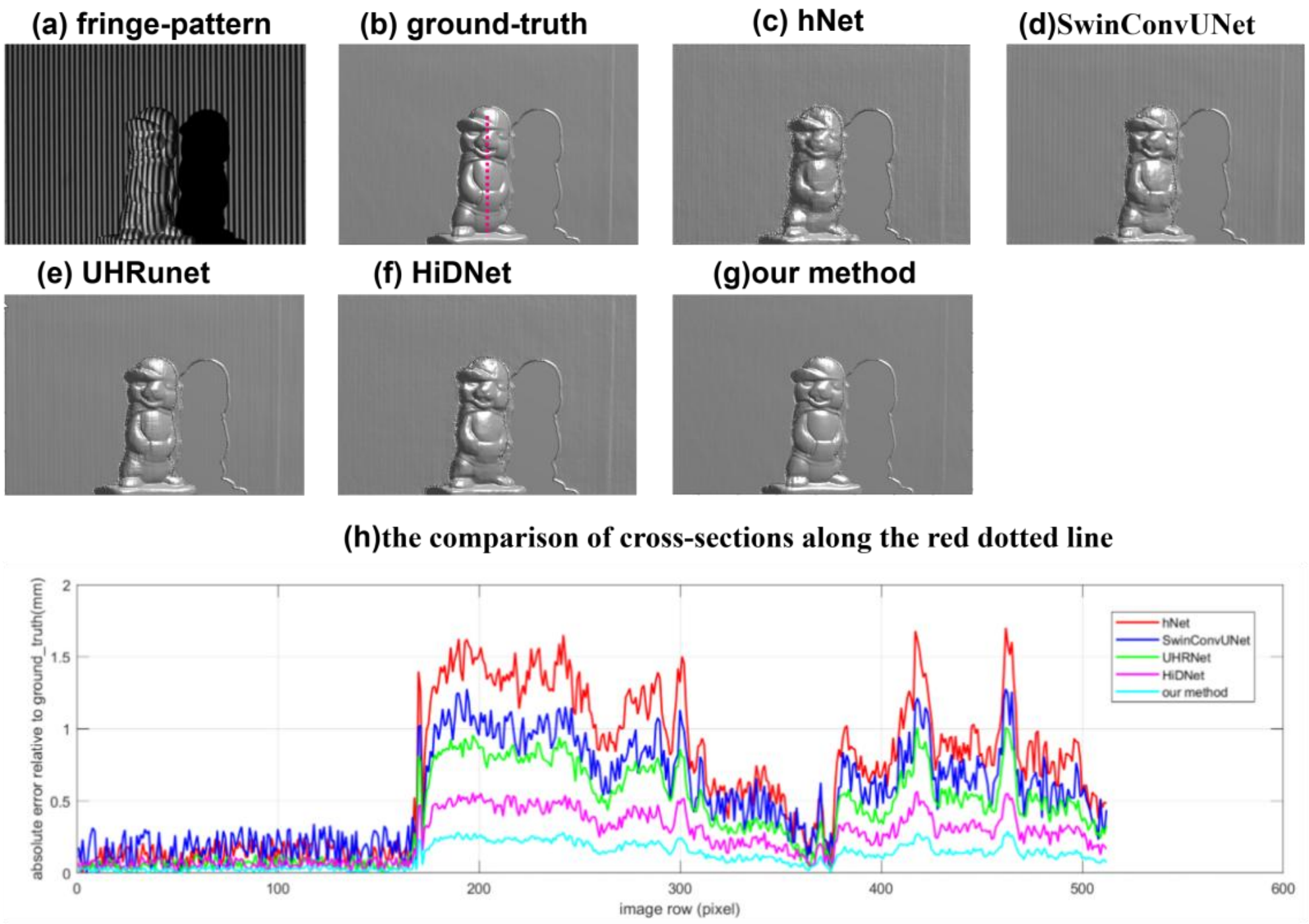

Fig. 8. Qualitative reconstruction comparison for a representative test sample: (a) input fringe pattern; (b) ground-truth height map; (c) – (g) height maps reconstructed using hNet, SwinConvUNet, UHRNet, HiDNet, and our method, respectively; and (h) absolute-error profiles sampled along the red dotted line in (b).

Figure 9 compares the reconstruction of two spatially isolated objects. For hNet, SwinConvUNet, UHRNet, HiDNet, and the proposed method, the RMSE values are 1.958, 1.695, 1.576, 1.358, and 0.549 mm, respectively, while the corresponding SSIM values are 0.9622, 0.9669, 0.9750, 0.9866, and 0.9983. The error maps reveal extensive residual errors on both objects for hNet and a pronounced error region on the left object for SwinConvUNet. These errors are reduced by UHRNet and HiDNet and reach their lowest level with the proposed method. The proposed method also preserves the separation between the objects and recovers their fine surface structures more faithfully.

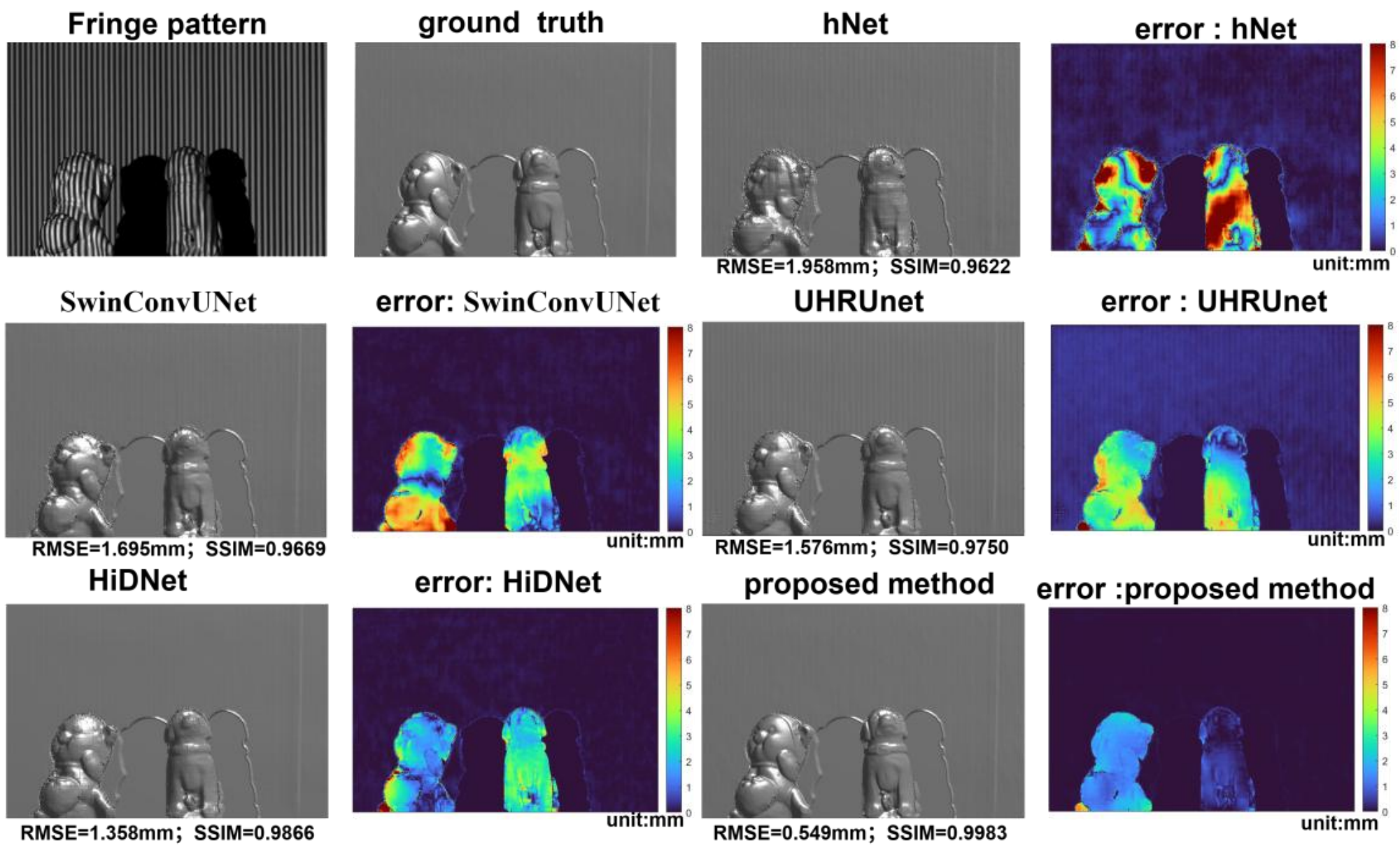

Fig. 9. Single-shot height reconstruction of two spatially isolated objects. The panels show the input fringe pattern, ground-truth height map, and reconstructed height and absolute-error maps obtained using hNet, SwinConvUNet, UHRNet, HiDNet, and the proposed method. All absolute-error maps use a common range of 0–8 mm.

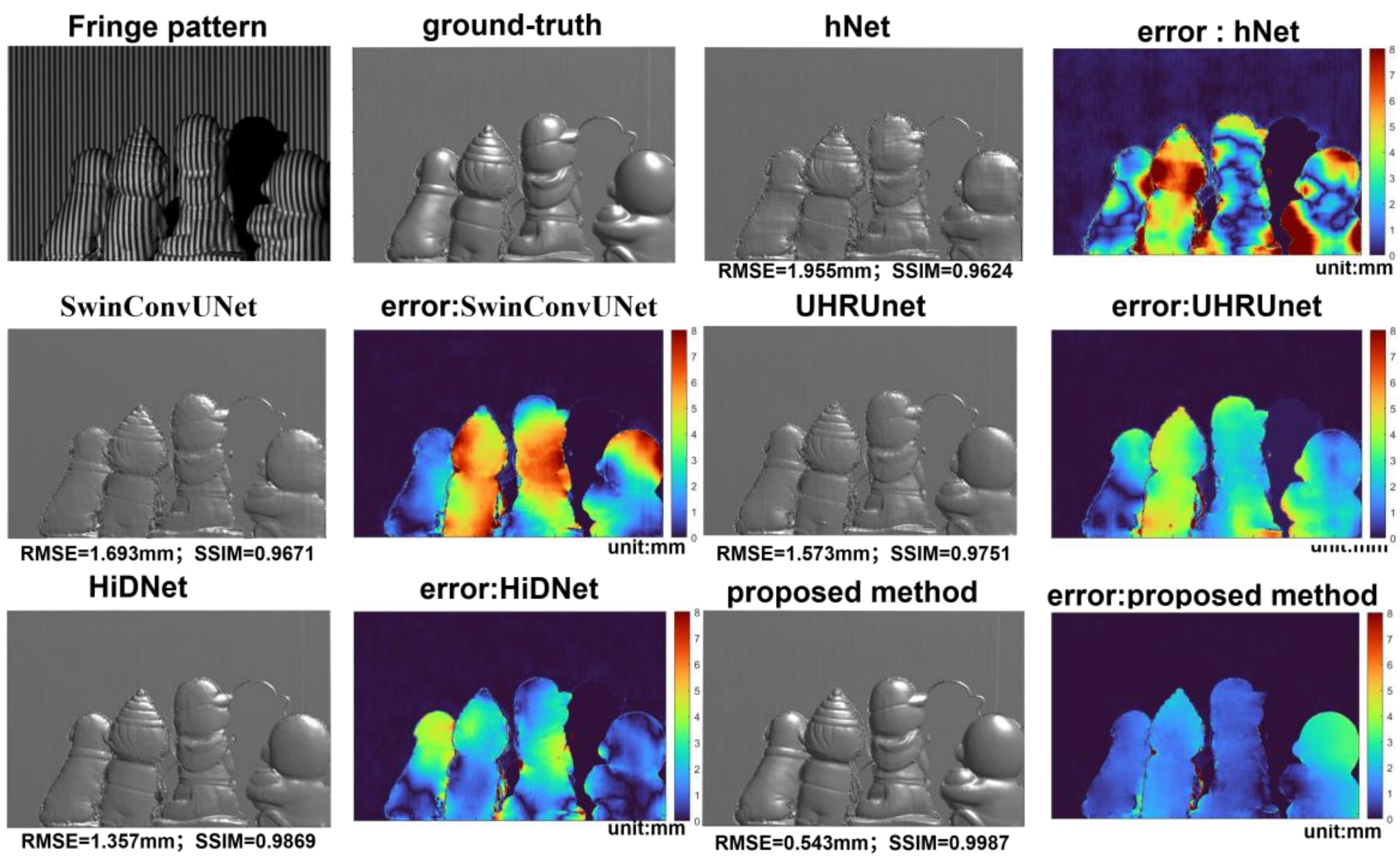

Fig. 10. Single-shot 3D reconstruction of multiple overlapping objects.

From left to right and top to bottom: the captured fringe pattern, the ground-truth height, and the height map followed by the absolute-error map of hNet, SwinConvUNet, UHRNet, HiDNet, and the proposed method. The error maps are in millimeters, All absolute-error maps use a common range of 0–8 mm

The third experiment considers a more complex scene containing multiple partially overlapping objects, with occlusions and pronounced depth discontinuities between adjacent surfaces. hNet, SwinConvUNet, UHRNet, HiDNet, and the proposed method were applied to the captured fringe pattern, with the results given in Fig. 10.

All methods recover the overall object arrangement, but the baseline reconstructions exhibit varying degrees of surface-detail loss and localized error. The proposed method more accurately preserves relative positions, object boundaries, and fine geometric structures. hNet, SwinConvUNet, UHRNet, and HiDNet reproduce the global shape, while the fine details become weaker, particularly in the hNet and SwinConvUNet results.

For hNet, SwinConvUNet, UHRNet, HiDNet, and the proposed method, the RMSE values are 1.955, 1.693, 1.573, 1.357, and 0.543 mm, respectively, and the corresponding SSIM values are 0.9624, 0.9671, 0.9751, 0.9869, and 0.9987. The proposed method reduces RMSE by 72.2% relative to hNet and by 60.0% relative to HiDNet. The absolute-error maps show the same trend, with the proposed method producing the smallest residual errors.

Overall, the quantitative and qualitative results consistently show that the proposed method provides more accurate surface-height reconstruction than the four representative baselines, particularly near object boundaries, fine surface structures, and depth discontinuities.

## 4. Discussions and conclusion

This study presents a hybrid Mamba–Transformer network for single-shot surface-height reconstruction from a deformed fringe pattern. The proposed architecture integrates selective state-space modeling with hierarchical window-based attention in a U-shaped encoder–decoder framework. At the intermediate-resolution $L_1$ stage, the proposed axial–dilated vision state-space (ADVSS) blocks combine four-directional selective scanning with local, dilated, and axial depth-wise convolutions. This design captures long-range spatial dependencies while retaining the directional and multiscale characteristics of fringe patterns without modifying the underlying selective-scan operator. Swin Transformer blocks are retained at the high-resolution encoder stage, bottleneck, and decoder to provide complementary window-based feature refinement.

The architecture ablations demonstrate that the allocation of VSS and Swin Transformer blocks across the network substantially affects reconstruction performance. Replacing the $L_1$ Swin Transformer blocks with VSS blocks improves all four evaluation metrics relative to the all-Swin baseline, whereas extending VSS blocks to the high-resolution $L_0$ stage does not provide an additional benefit. Replacing the $L_1$ VSS blocks with ADVSS blocks produces the best overall configuration. Furthermore, removing the axial, dilated, or post-scan local residual component degrades performance, supporting the complementary roles of directional encoding, enlarged-context extraction, and local-detail preservation. The input-channel experiments also show that the five-channel representation does not improve the VSS- or ADVSS-based models under the present training protocol. The proposed single-channel configuration can therefore recover surface geometry directly from one deformed fringe image without requiring reference-fringe or coordinate channels.

Under the unified evaluation protocol, the proposed method outperforms hNet, SwinConvUNet, UHRNet, and HiDNet in MAE, RMSE, AbsRel, and SSIM. It achieves an MAE of 0.195 mm, an RMSE of 0.548 mm, an AbsRel of 0.043, and an SSIM of 0.9985. Compared with UHRNet and HiDNet, its RMSE is reduced by 65.2% and 59.6%, respectively. These findings indicate that combining selective scanning with hierarchical window-based feature processing is effective for single-shot fringe-based surface reconstruction.

The improvement in accuracy is accompanied by increased computational cost relative to lightweight CNN-based models. The proposed network contains 30.225 M parameters and requires 0.0246 s per image on an RTX 4090, compared with 8.631 M parameters and 0.0092 s for hNet. Although its latency is only moderately higher than those of SwinConvUNet and UHRNet, further optimization is required for resource-constrained platforms and high-speed measurement applications.

Several limitations should be acknowledged. First, the experiments were conducted using data acquired under a fixed imaging configuration and fringe spatial frequency. Generalization across different cameras, projection systems, working distances, fringe densities, and object materials therefore remains to be established. Second, highly reflective or non-Lambertian surfaces, shadows, steep slopes, and regions with low fringe modulation may still cause locally elevated errors.

Future work will evaluate the proposed framework on more diverse real-world datasets and across different measurement systems. Physics-informed constraints, uncertainty estimation, and degraded-fringe augmentation will be investigated to improve reliability in low-quality fringe regions. Lightweight architectural variants, model compression, and hardware-oriented acceleration will also be explored to support real-time implementation.

In conclusion, the results demonstrate that integrating ADVSS-based selective state-space modeling with hierarchical window-based attention provides an effective approach to accurate single-shot fringe-projection surface reconstruction.

**CRediT authorship contribution statement**

Xinjie Zhou: Writing- original draft, Visualization, Software,Methodology, Investigation, Conceptualization,Formal analysis,Data curation;

XiaoLei Li: Writing-review & editing,Software, Project administration, Methodology, Conceptualization,Formal analysis, Data curation.

**Disclosures**

The authors declare that they have no known competing financial interests or personal relationships that could have appeared to influence the work reported in this paper.

**Data availability**

The source code, trained models, and processed dataset will be made publicly available after the manuscript is accepted.

The release will be published at:https://github.com/Zhou-del-1/Zhou-del-1.github.io

The datasets are from reference [8] (https://zenodo.org/records/15872069).